\documentclass[10pt,twocolumn,letterpaper]{article}

\usepackage{iccv}
\usepackage{times}
\usepackage{epsfig}
\usepackage{graphicx}
\usepackage{amsmath}
\usepackage{amssymb}
\usepackage{multirow}
\usepackage{enumitem}

\usepackage[pagebackref=true,breaklinks=true,letterpaper=true,colorlinks,bookmarks=false]{hyperref}

\iccvfinalcopy 

\def\iccvPaperID{7332} 

\ificcvfinal\fi

\begin{document}

\title{Breaking Down the Task: A Unit-Grained Hybrid Training Framework for Vision and Language Decision Making}

\author{Ruipu Luo\\
Fudan University\\
Shanghai, China\\
{\tt\small rpluo21@m.fudan.edu.cn}
\and
Jiwen Zhang\\
Fudan University\\
Shanghai, China\\
{\tt\small  jiwenzhang21@m.fudan.edu.cn}
\and
Zhongyu Wei$^{\dag}$\\
Fudan University\\
Shanghai, China\\
{\tt\small zywei@fudan.edu.cn}
}

\maketitle
\footnotetext[1]{$^{\dag}$~Corresponding author.}
\ificcvfinal\thispagestyle{empty}\fi

\begin{abstract}
    Vision language decision making (VLDM) is a challenging multimodal task. The agent have to understand complex human instructions and complete compositional tasks involving environment navigation and object manipulation. However, the long action sequences involved in VLDM make the task difficult to learn. From an environment perspective, we find that task episodes can be divided into fine-grained \textit{units}, each containing a navigation phase and an interaction phase. Since the environment within a unit stays unchanged, we propose a novel hybrid-training framework that enables active exploration in the environment and reduces the exposure bias. Such framework leverages the unit-grained configurations and is model-agnostic. Specifically, we design a Unit-Transformer (UT) with an intrinsic recurrent state that maintains a unit-scale cross-modal memory. Through extensive experiments on the TEACH benchmark, we demonstrate that our proposed framework outperforms existing state-of-the-art methods in terms of all evaluation metrics. Overall, our work introduces a novel approach to tackling the VLDM task by breaking it down into smaller, manageable units and utilizing a hybrid-training framework. By doing so, we provide a more flexible and effective solution for multimodal decision making.

\end{abstract}

\section{Introduction}
    Recent years have witneessed an increasing number of embodied agents in our daily life, such as food delivery robot in the restaurant and sweeping robot designed for house-keeping. These robot assistants take natural language as input and interact with the environment accordingly. In order to enhance the capability of language-driven embodied agents, various Vision and Language Decision Making (VLDM) tasks and benchmarks have been proposed~\cite{shridhar2020alfred,padmakumar2021teach}, where the agent is required to complete compositional tasks under human instructions. During the process, they need to execute a sequence of actions for \textit{navigation} and \textit{object interaction}. For example, to complete the ``slicing the bread'' task, the agent needs to navigate towards the bread, pickup the bread, then put the bread on the countertop and finally execute the slicing action. 
    \begin{figure}
        \centering
        \includegraphics[width = 0.9\linewidth]{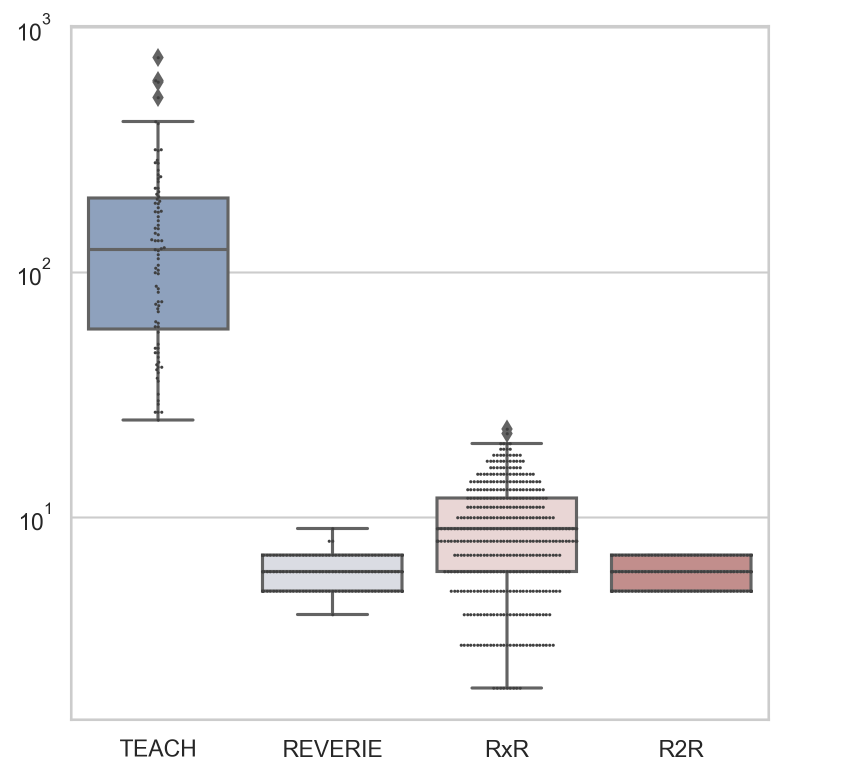}
        \caption{\textbf{Boxplots of logarithmic action sequence lengths} for different datasets. We compare the VLDM dataset TEACH\cite{padmakumar2021teach} with three VLN datasets including RXR\cite{ku2020room}, REVERIE\cite{qi2020reverie}, and R2R\cite{anderson2018vision}. The average length of the action sequence in TEACH is 157, average length of the action sequence in RXR is 9, and average length of the action sequence in R2R and REVERIE is 6.}
        \label{fig:intro}
    \end{figure}
    
    \begin{figure*}[t]
        \centering
        \setlength{\abovecaptionskip}{0.1cm}
        \includegraphics[width=\linewidth]{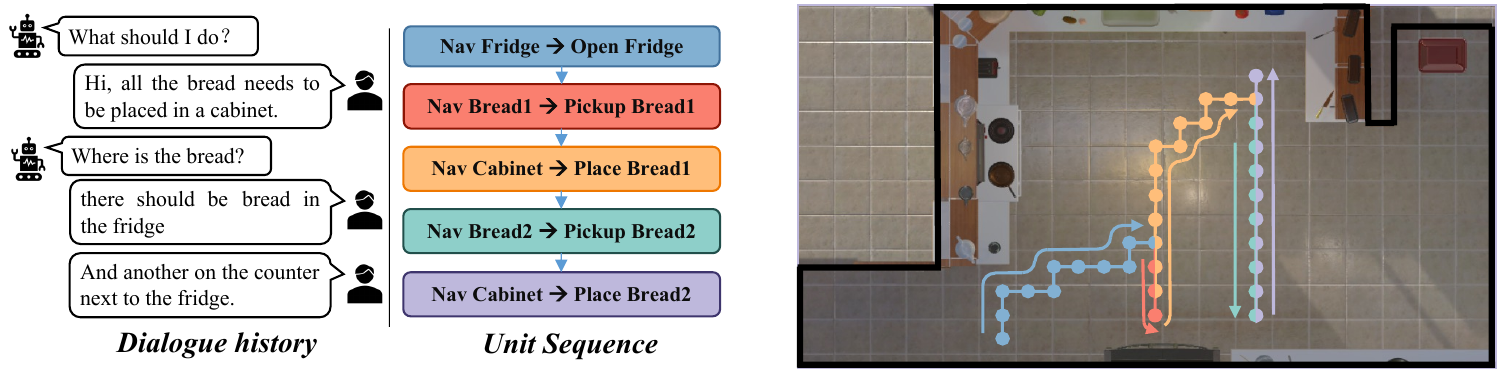}
        \caption{\textbf{An unit grained example of EDH instance.} The task dialogue history is on the left and the whole navigation path is on the right. Different color indicates different unit. Each unit starts with a navigation phase and is ended by an object interaction, i.e. the orange path represents that an agent navigates to cabinet and places the bread which is picked up in last unit.}
        \label{task_redefine}
        \vspace{0em}
    \end{figure*}
    
    VLDM tasks usually involves hundreds of actions to complete a compositional task. As shown in Figure \ref{fig:intro}, the complexity of the VLDM task is much greater than that of the VLN, causing low-efficiency in optimizing the model with human demonstration action sequence. Existing methods ~\cite{pashevich2021episodic,shridhar2020alfred} feed the entire action sequence into the model for direct training, and the learning effect for long sequence is insignificant. We observe that VLDM tasks can be decomposed into multiple sequential subtasks based on the type of ending action. Each subtask contains a navagation phase and an object interaction phase. The agent needs to navigate to target object before interacting with the object. We design a segmentation framework named \textit{unit-grained segmentation}, which divides long action sequences of original data episode into multiple instances called \textit{units}. We use the segmented short sequences for unit-grained training. Figure \ref{task_redefine} shows a segmentation example. 
    
    To train a embodied agent, most existing method ~\cite{pashevich2021episodic,shridhar2020alfred,min2021film} following the fashion of behavior cloning, where the model takes the human demonstration of the previous step as input and predicts the action for current step. However, the human demonstration action is not available during inference. This results in the problem of \textit{exposure bias}~\cite{ross2011reduction} in sequence modeling. Student forcing is hired to eliminate the gap between training and inference~\cite{anderson2018vision,fried2018speaker,ma2019self}, where
    predicted actions are executed to get new environment observations and fed to the model for next action prediction. 
    However, object interaction actions in VLDM tasks make student forcing strategy not directly applicable. Our observation is that the environment state is changed only when the agent manipulates objects. Therefore, we split the episode into several units based on object interactions and build an offline environment for each unit-grained instance such that the agent can move freely according to its own prediction during training. Offline environment of each unit ensures accurate self-centered observation for any agent actions. Moreover, we propose a hybrid forcing training strategy that allows agent to perform student forcing training first, and then conduct teacher forcing training after reaching the maximum number of trial steps.
    
    In summary, our main contributions are as follows:
    \begin{itemize}[topsep=1pt]
        \setlength{\itemsep}{1pt}
        \setlength{\parsep}{2pt}
        \setlength{\parskip}{0pt}
        \item We reconstruct original VLDM data to unit-grained instances and build an offline environment that enables efficient free exploration during training.
        \item We propose to train embodied agent via a novel hybrid-training framework that combines the advantages of teacher-foring and student forcing strategy.
        \item Under the unit-grained task configurations, we design an iterative model called Unit Transformer (UT) with a unit-scale intrinsic recurrent state.
    \end{itemize}
    Experiments conducted on TEACH dataset indicate our unit-grained settings can replace the original episodic settings to achieve state-of-the-art results. Ablation studies demonstrate the effectiveness of proposed hybrid training framework and model architecture.
    

\section{Related Works}
    \paragraph{Vision and Language Decision.} 
    Vision language Decision Making (VLDM) includes both navigation-only tasks (VLN) and navigation plus object interaction tasks. VLN tasks only require the agent to move to the target position according to the instructions. 
    This task has been extensively studied in recent years \cite{anderson2018vision,JacobKrantz2020BeyondTN,ArunBalajeeVasudevan2021Talk2NavLV}.
    Some benchmarks \cite{anderson2018vision,jain2019stay,ku2020room,yan2019cross,YiWu2018BuildingGA,AbhishekDas2017EmbodiedQA} take fine-grained language instructions that describe each step during navigation as input,
    while other benchmarks use coarse instructions \cite{TaChungChi2019JustAI,KhanhNguyen2019VisionBasedNW,KhanhNguyen2019HelpAV} or dialogue with humans \cite{JesseThomason2019VisionandDialogN,ShurjoBanerjee2020TheRB,XiaofengGao2022DialFREDDA,AnjaliNarayanChen2019CollaborativeDI}.
    Unlike navigation only tasks, the VLDM task \cite{DipendraMisra2018MappingIT,shridhar2020alfred,padmakumar2021teach} is more general for embodied AI. The agent has to not only navigate towards the target location and but also do multiple object operations \cite{shridhar2020alfred}, such as ``Preparing Breakfast". 
    Recent works on VLDM tasks do not distinguish the navigation and interaction actions, ignoring the environment changes caused by object operations. The agent may predict to operate an object even though it does not see any object in the current view. To tackle this problem, we explicitly divide the navigation and interaction phases of each episode in this paper.

    \paragraph{Teacher \& Student Forcing Training Strategy.}
    For sequence generation, teacher and student forcing training are commonly used and closely related \cite{haidar2019textkd,rashid2020unsupervised,lee2021contrastive,fedus2018maskgan}. Teacher forcing uses the ground truth of the previous step as the current input, whereas student forcing uses the prediction of the previous step. Teacher forcing strategy can correct the predictions of the model during training and avoid further amplification of errors \cite{pashevich2021episodic,shridhar2020alfred}, but it also 
    causes exposure bias and over correction. Under navigation only settings \cite{fried2018speaker,wang2019reinforced,tan2019learning,ma2019self}, student forcing is applied to explore the environment and better generalize in unseen scenarios. Some studies even use the DAgger-style student forcing strategy \cite{StephaneRoss2010ARO} to sample an action. We combine the advantages of above-mentioned two strategies by proposing a novel hybrid forcing training framework. Such framework takes into account not only the learning stability during training but also the generalization performance at inference time. 

    
    \paragraph{Multimodal Pretraining with Transformers}  
    In recent years, since transformers are applied to extract visual features, such as ViT\cite{dosovitskiyimage}, transformer structure is widely used in multimodal representation aera. Transformers have shown significant progress in vision and language tasks, achieving state-of-the-art performance in downstream tasks such as visual language question answering \cite{khare2021mmbert,biten2022latr}, image captioning\cite{yu2019multimodal}, etc. 
    Both one-stream \cite{su2019vl,li2020unicoder} and two-stream \cite{tan2019lxmert,lu2019vilbert,radford2021learning} architecture can perform good feature fusion between multiple modalities. Some studies \cite{hong2021vln,qiao2022hop} introduced multimodal transformers into VLN tasks. VLN-BERT \cite{hong2021vln} equips the BERT model with a recurrent function mechanism that maintains cross-modal state information for the agent. HOP \cite{qiao2022hop} considers historical information and sequential relations, and designs multiple pre-training tasks to adapt to the specificity of VLN. Inspired by these works, we propose a one-stream multimodal transformer model, namely Unit Transformer, that fuses text, images and actions and incorporates a memory state vector to record historical information.

    \begin{figure}
        \raggedleft
        \includegraphics[width=\linewidth]{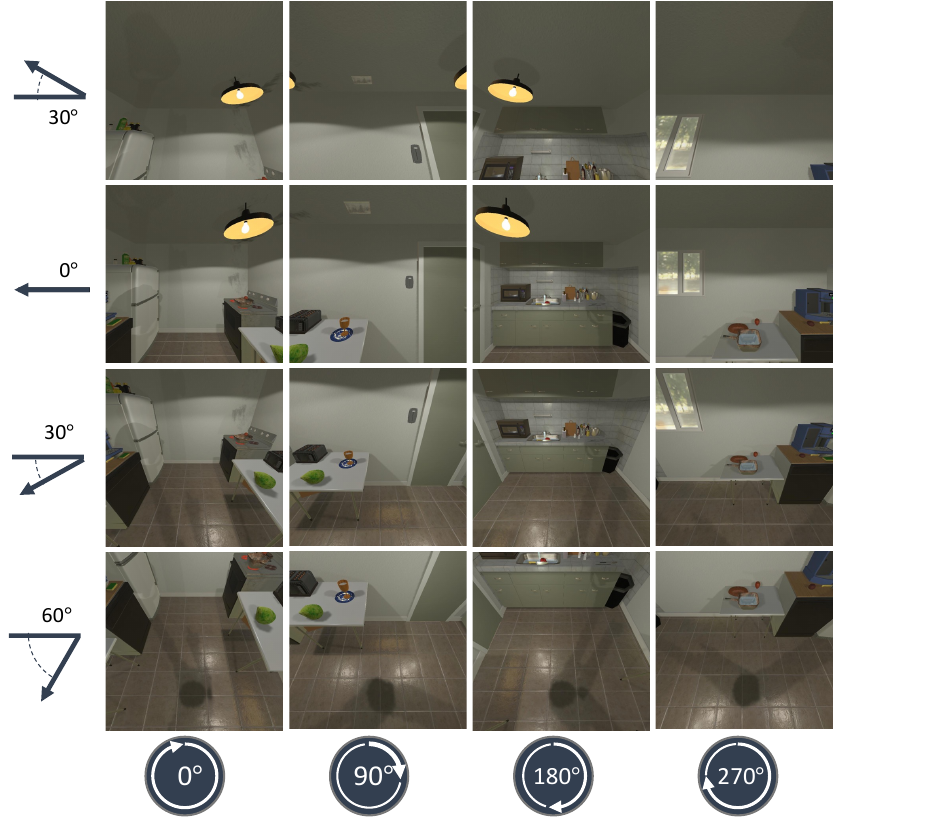}
        \caption{\textbf{Panorama of single point sampling in unit offline environment.} Rows are the observations of the vertical rotation angle of the agent from 30 degrees looking up to 60 degrees looking down. Columns are the observations of the agent rotating clockwise at intervals of 90 degrees.}
        \label{offline_env}
    \end{figure}

    \begin{table}[t]
    \small
    \begin{center}
    \begin{tabular}{|c|ccc|}
    \hline
    \multicolumn{1}{|c|}{}    & \multicolumn{3}{c|}{\textbf{EDH Instance}}                     \\ \hline
    \textbf{Split}                    & \multicolumn{1}{c|}{\textit{train}} & \multicolumn{1}{c|}{\textit{val\_seen}} & \textit{val\_unseen} \\ \hline
    \textbf{\#}           & \multicolumn{1}{c|}{5475}  & \multicolumn{1}{c|}{608}   & 2175  \\ \hline
    \textbf{Action Length}    & \multicolumn{1}{c|}{77.31} & \multicolumn{1}{c|}{73.46} & 75.43 \\ \hline
    \textbf{\# of Dialogue Turns} & \multicolumn{1}{c|}{11.14}          & \multicolumn{1}{c|}{10.89}              & 9.95                 \\ \hline
    \textbf{Dialogue Lengths} & \multicolumn{1}{c|}{22.37} & \multicolumn{1}{c|}{21.53} & 19.82 \\ \hline
    \end{tabular}%
    \end{center}
    \caption{\textbf{Statistics for EDH instances.} Even if the teach session is divided into EDH instances, the average action length is still greater than 70, and whole task is still too complicated.}
    \label{EDH instance table}
    \end{table}
    
    \begin{table}[t]
    \small
    \begin{center}
    \begin{tabular}{|c|ccc|}
    \hline
                                      & \multicolumn{3}{c|}{\textbf{Unit Instance}}                     \\ \hline
    \textbf{Split} & \multicolumn{1}{c|}{\textit{train}} & \multicolumn{1}{c|}{\textit{val\_seen}} & \textit{val\_unseen} \\ \hline
    \textbf{\#}                   & \multicolumn{1}{c|}{27920} & \multicolumn{1}{c|}{3380}  & 12741 \\ \hline
    \textbf{Action Length}            & \multicolumn{1}{c|}{5.22}  & \multicolumn{1}{c|}{5.14}  & 4.85  \\ \hline
    \textbf{\# of Dialogue Turns} & \multicolumn{1}{c|}{5.31}  & \multicolumn{1}{c|}{4.93}  & 5.04  \\ \hline
    \textbf{Dialogue Lengths}         & \multicolumn{1}{c|}{33.27} & \multicolumn{1}{c|}{31.49} & 30.26 \\ \hline
    \end{tabular}%
    \end{center}
    \caption{\textbf{Statistics for Unit-grained instances.} After segmenting TEACH session by our method, we get 27920 data instances on training set. The unit instance has an average action sequence length of about 5. It can be seen that the data complexity after unit-grained instance is reduced.}
    \label{unit instance table}
    \end{table}

    \begin{figure*}[ht]
    \centering
    \setlength{\abovecaptionskip}{0.1cm}
    \includegraphics[width=\linewidth]{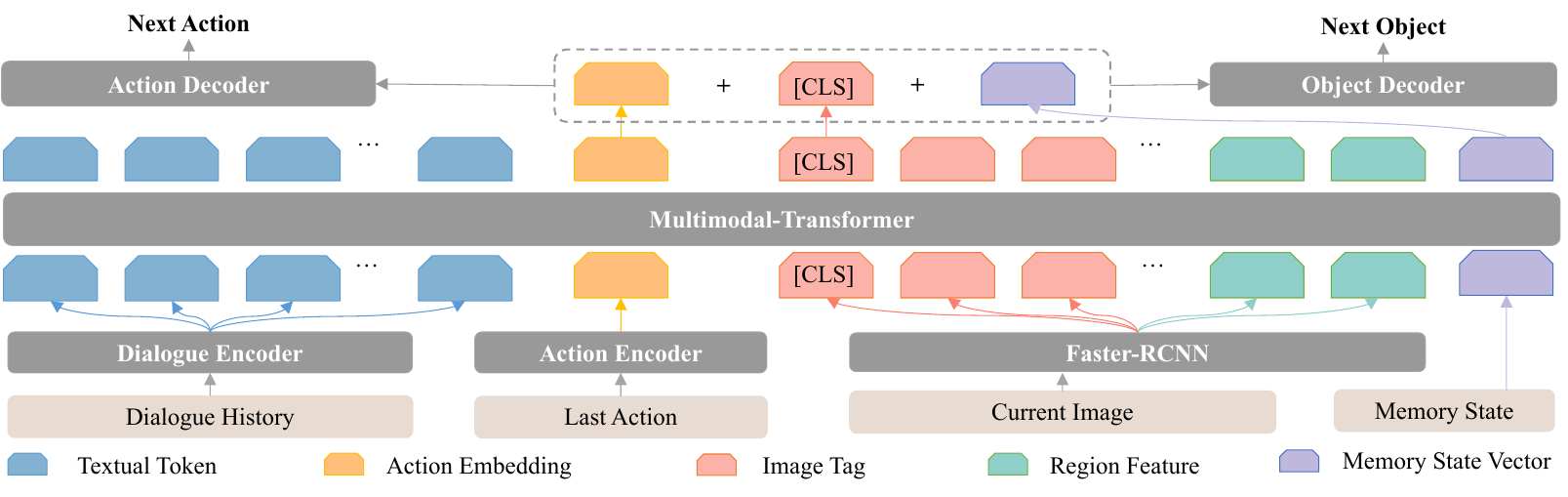}
    \caption{\textbf{Structure of Unit Transformer.} Blue, yellow, red, green, and purple represent information about textual, action, object tags, regional features, and memory states, repstively. After passed through their corresponding encoders, vectors are fed into a two-layer transformer to fuse features and obtain the final classification distributions.}
    \label{fig:UT}
    \vspace{0em}
    \end{figure*}

\section{Data Reconstruction}
    In this section, we firstly introduce the setup of a typical VLDM task presented in TEACH \cite{padmakumar2021teach}. Then we introduce our fine-grained data reconstruction method called unit segmentation. According to the segmented unit instances, we construct a corresponding offline environment for each unit, supporting free exploration during offline training.

    \subsection{Preliminary of EDH Benchmark}
    
    TEACH presents a benchmark named Execution from Dialogue History (EDH), a typical VLDM task. This benchmark is collected by online simulator AI2THOR, containing 98 indoor scenes. Each TEACH session includes a complete process of agent (called Follower) performs a household task (like ``Put All X into Y") in which the instructions are given in the form of dialogue by another agent (called Commander). Each session includes an initial state $S_i$ and an final state $S_f$ (the state includes both agent state and environment state), dialogue information $D$, and the sequence of actions $A = \{a_1, a_2, \dots ,a_n\}$ performed by the agent. To determine the task completeness, we only have to check whether the final state is consistent.
    
    TEACH sessions are segmented into EDH instance. One EDH instance is denoted as a tuple $(D_H,A_H,A_F,S^E_i,S^E_f)$ where $D_H$ is the dialogue history, $A_H$ is the action history, $A_F$ is the future action, $S^E_i$ is the EDH initial state, $S^E_f$ is the EDH final state. The action sequence $A$ consists of the sequence of action history and future action, denoted as $[A_H,A_F]$. There are 8 object interaction actions (Pick up, Place, Poor, Slice, Open, Close, Toggle On, Toggle Off) and 8 navigation actions (Forward, Backward, Pan Left/Right, Turn Left/Right, Look Up/Down) in the action space. The agent learns to complete the task following the dialogue history and action history. At each time step, the agent executes one of the above mentioned 16 actions. Table \ref{EDH instance table} shows the statistics of EDH instances.

    \subsection{Unit-grained Instance Segmentation}
    We observe that the task in original TEACH session is too complicated, i.e. the length of the action sequence is too long. In TEACH benchmark, authors divide long session into EDH instances to reduce the difficulty of the task. However, even after EDH segmentation, the average length of EDH action sequence still reaches hundreds of steps.

    We propose a unit-grained segmentation method. We observe that task in a TEACH session consists of a series of stages that agent need to interact with the environment. An example is shown in Figure \ref{task_redefine}, where the commander asks the follower to place all bread into the cabinet with a hint that one bread is in fringe and another is on the counter next to fringe. Such a session can naturally be segmented into 5 high-level instances according to the execution of interactive actions: \textit{Navigate and Open Fringe}, \textit{Navigate and Pickup Bread1}, \textit{Navigate Cabinet and Place bread1}, \textit{Navigate and Pickup Bread2}, \textit{Navigate Cabinet and Place Bread2}. Each high-level instance, denoted as \textbf{\textit{unit}}, contains a navigation phase and a interaction phase. 
    We represent a unit instance as a tuple $(U_l, U_n, S^u_i, D^u_H, A_u)$, where $U_l$ and $U_n$ indicate last and next unit. $S^u_i$ is initial state of current unit. $D^u_H$ represents all dialogue history that occurred before current unit. Agent let $S^u_i$ and $D^u_H$ as input and outputs an action sequence $A_u$ in current unit. The statistics of unit instances are shown in Table \ref{unit instance table}.

    \subsection{Offline Environgment Building \label{offline data building}}
    In pure-navigation VLDM tasks, agent can move freely in the environment during the offline training process by student forcing training strategy. For example, if agent is trained by R2R dataset, it can navigate freely by following a self-predicted path during offline training. However, since TEACH session is collected in online AI2THOR simulator, and action performed by the agent involves object manipulations that leads to state changing of environment. Therefore, the agent can only following the ground truth path during offline training. This increases the inconsistency between training and inference.

    After unit segmentation, the state of environment in each unit retain the same. This inspires us to collect panoramic images of all points that agent can reach in the environment. The panorama of each point includes total 16 pictures in the horizontal direction of 0 degrees, 90 degrees, 180 degrees, 270 degrees and vertical downward directions of -30 degrees, 0 degrees, 30 degrees and 60 degrees. These panoramas enable the agent to get the correct egocentric picture after performing any action in the current unit. Such an offline environment allows the agent to actively explore the environment during training. An example of panorama collection at a single point is shown in Figure \ref{offline_env}.

\section{Methodology}


In this section, we introduce the unit transformer model and hybrid forcing training strategy. The unit transformer combines text, image, and action information to accurately predict the agent's next action and its corresponding object. To facilitate unit segmentation, we have incorporated a memory state vector that implicitly captures the step state of the current unit. The structure of unit transformer is shown in Figure \ref{fig:UT}. Furthermore, we propose a hybrid forcing training strategy that leverages both student and teacher forcing training methods to enhance the performance of our unit transformer model.

    \subsection{Unit Transformer \label{unit-model}}
    Under our unit-grained instance, when the agent makes a decision at time $t$, the information it can obtain are the instruction dialogue history before current unit, the action performed in the previous step, and the egocentric image of the current location. Agent need to take several navigation action and execute one interaction action in a unit. In order for the agent to remember the process history of current unit, the agent will also obtain a memory state. Therefore, the input of the model should contain instruction dialogue, last action, current egocentric image and memory state vector, which denoted as a tuple $(I, action_{t-1}, img_{t}, s_{t-1})$.
    
    \paragraph{Multi-modal Feature Extraction} Since the name of the action itself includes some semantic information (such as “Turn Left” will let agent more focus on left side of image) , we also use a text encoder to obtain the action representation. We concatenate all sentences in dialogue as one sentences. In practice, we use a trainable embedding matrix as a text encoder. The dialogue embedding and action embedding can be obtained as follow:
    \begin{equation}
    \begin{aligned}
    \{I_1,...,I_n\} &= \operatorname{TextEncoder}(I)\\
    a_{t-1} &= \operatorname{TextEncoder}(action_{t-1}),
    \end{aligned}
    \end{equation}
    where $n$ is length of sentences. Since the relative positions of objects (such as a cup on a table) are often used when describing action instructions, it is indispensable to obtain regional features of objects. We adopt an object detection model Faster R-CNN as the region feature extractor. It takes a egocentric image as input, and outputs labels $\boldsymbol{l}_t$, bounding boxes $\boldsymbol{b}_t$, and region features $\boldsymbol{r}_t$. The formula is:
    \begin{equation}
    \boldsymbol{l_t,b_t,r_t} = \operatorname{FasterRCNN}(img_t)\\
    \end{equation}
    where $m$ is the number of detected objects and $\boldsymbol{l}_t=\{l_t^1,...,l_t^m\}$, $\boldsymbol{b}_t=\{b_t^1,...,b_t^m\}$, $\boldsymbol{r}_t=\{r_t^1,...,r_t^m\}$. In order to make better use of the extracted object information, we concatenate the 4 coordinates, width and height of the bounding box to the back of the regional features.The concatenated regional features will feed into one layer MLP to unify the feature dimensions. The final object labels and regional features can be calculated as follows:
    \begin{equation}
    \begin{aligned}
        \{\hat{l}_t^1,...,\hat{l}_t^m\} &= \operatorname{TextEncoder}(\boldsymbol{l}_t)\\
        \{\hat{r}_t^1,...,\hat{r}_t^m\} &= \operatorname{MLP}([\boldsymbol{r}_t;\boldsymbol{b}_t])
    \end{aligned}
    \end{equation}
    
    \paragraph{Feature Fusion and Decoding} We add a ``[CLS]" label in front of object labels $\{\hat{l}_t^1,...,\hat{l}_t^m\}$ to fuse the information of all regional features and object labels. We concatenate dialogue feature, last action feature, object tag feature, region feature and the memory state vector, and then input the two-layer multi-modal transformer to obtain the fusion representation of each modality. Then we concatenate the vector of actions $\tilde{a_{t-1}}$, ``[CLS]" label $\tilde{l_t^0}$ and memory states $s_t$ to predict the next action and object, mathematically expressed as follows:
    \begin{equation}
        \begin{aligned}
            a_t &= \operatorname{ActionDecoder}([\tilde{a}_{t-1};\tilde{l_t^0};s_t])\\
            o_t &= \operatorname{ObjectDecoder}([\tilde{a}_{t-1};\tilde{l_t^0};s_t])
        \end{aligned}
    \end{equation}

    \begin{figure}[t]
        \centering
        \includegraphics[width=\linewidth]{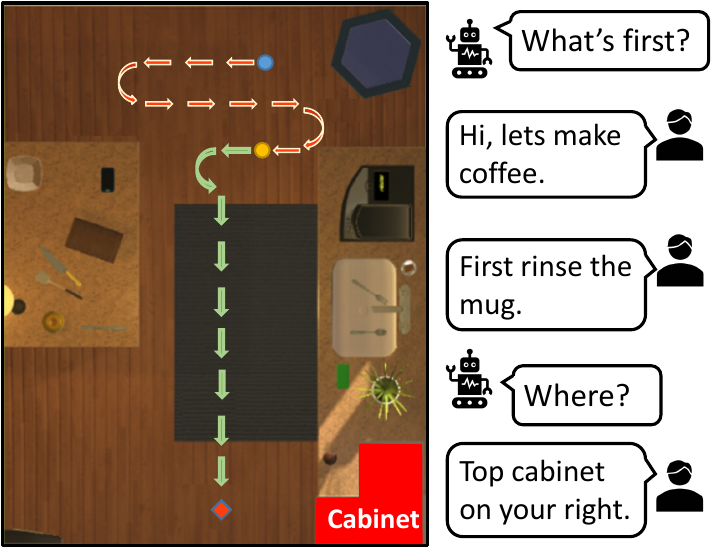}
        \caption{\textbf{Illustration of hybrid forcing training in a single unit.} The Commander send an instruction in the unit: "Make coffee", and give the location of coffee mug which is in the cabinet (red aera in left picture) on the right. Left picture shows the hybrid training process in this unit, with the blue point representing the initial position and the red point indicating the target position. Agent starts with student forcing training, in which it moves freely without stopping until it reaches the maximum number of steps preset in advance. The orange path in the picture represents the trajectory of the agent during this stage. The agent then switches to teacher forcing training, during which it follows the green path, which is the optimal path from the yellow point to the target point. The agent performs teacher forcing training according to this path.}
        \label{hybrid_train}
    \end{figure}
    \subsection{Hybrid Forcing Training Strategy}

    We propose a hybrid training strategy that combines both teacher forcing and student forcing strategy. Teacher forcing is a method for quickly and efficiently training recurrent models that use the ground truth from a prior time step as input, while student forcing use model output from prior time step as input. The Student forcing training strategy is widely used in pure navigation VLDM tasks for offline training. However, when the agent need to interacts with objects in the environment, the environment will change dynamically, and the student forcing training strategy is non-trival. Under our novel unit segmentation data setting, the student training strategy can be applied to offline training process, because (1) agent can obtain correct image observation through our offline environment and (2) state of environment is unchanged in one unit. The hybrid forcing training process in a single unit is shown in the Figure \ref{hybrid_train}.
    
    \subsubsection{Single Step Inference\label{single step}}
    In each unit instance, model input is denoted as a tuple $(a_{1:T}^{*}, v_{1:T}^{*}, D^u_H, s_0, POS_i, POS_T)$, where $a_{1:T}^{*}, v_{1:T}^{*}$ represent ground truth action and image observation from $time\ 1$ to $time\ T$. $POS_0$ and $POS_T$ are agent initial position and target position in environment. $POS_t$ is a 4 dimension vector $(x_t,y_t,hor_t,ver_t)$, where $x_t,y_t$ are point coordinates, $hor_t$ and $ver_t$ denote horizontal and vertical rotation degree as mentioned in section \ref{offline data building}. Agent aims to move from $ POS_0$ to $POS_T$ in current unit, and take a interaction action at time step $T$. The model calculation for each time step is:
    \begin{equation}
        \hat{s}_{t}, \hat{a}_{t} = \operatorname{UT}(D_T,a_{t-1},v_{t},s_{t-1})
    \end{equation}
    When using the teacher forcing strategy, the inputs $a_{t-1}$ and $v_{t}$ come from the ground truth action and image observation. While applying student forcing strategy, the input $a_{t-1}$ comes from the action output by the agent prediction in previous step. During student forcing training stage, we restrict the action output to be navigable only, and agent is able to take this action and obtain current position $POS_t$ in offline environment. Current image observation $v_{t}$ is obtained by inputting current position $POS_t$ to offline environment, denoted as $v_{t} = ENV(POS_t)$. Single step inference is represented as follows:
    \begin{equation}
    \label{eq6}
    a_{t-1},v_{t}=\left\{
    \begin{aligned}
    a_{t-1}^{*}& ,v_{t}^{*} ,  & \text{if teacher forcing}, \\
    \hat{a}_{t-1}& , \operatorname{ENV}(POS_t) , & \text{if student forcing}.
    \end{aligned}
    \right.
    \end{equation}
    
    \subsubsection{Hybrid Forcing Training Process}
    Hybrid Forcing Training Process includes two stages (student forcing and teacher forcing) during offline training. In first stage, agent predicts an action $a_t$ in each step using student forcing way that introduced in section\ref{single step}. We generate an optimal path from current position$POS_t$ to target position $POS_T$, which can obtain agent's next optimal position $POS^{*}_{t+1}$. We then compare the current position with the next optimal position to generate the ground truth action $a^*_t$, an illustration is shown in Figure \ref{hybrid_train}. Using this generated $a^*_t$ we compute this step's loss by using cross entropy loss function denoted as $loss^t_s$. Finally, Agent execute predicted $a_t$ and move to position $POS_{t+1}$ in offline environment. To prevent agent wandering endlessly in the same place, we limit the maximum number of steps as 5 plus the length of ground truth path during student forcing stage. 
    
    In teacher forcing stage, if the agent can not navigate to target position when the maximum step number is reached, agent is at wrong position and get inaccurate memory state to perform the last interaction action of the unit. We generate an optimal path from current position to target position, and obtain an optimal action sequence. Agent do action prediction and loss computation in teacher forcing way follow the optimal action sequence. For initialization of $a_0$ and $s_0$, we use last action and last state of last unit. If unit is first unit among whole unit segmentation, representation of token ``[Start]" and ``[CLS]" are used to initialize $a_0$ and $s_0$.

\section{Experiment}

    \begin{figure}[t]
    \centering
    \setlength{\abovecaptionskip}{0.1cm}
    \includegraphics[width=0.95\linewidth]{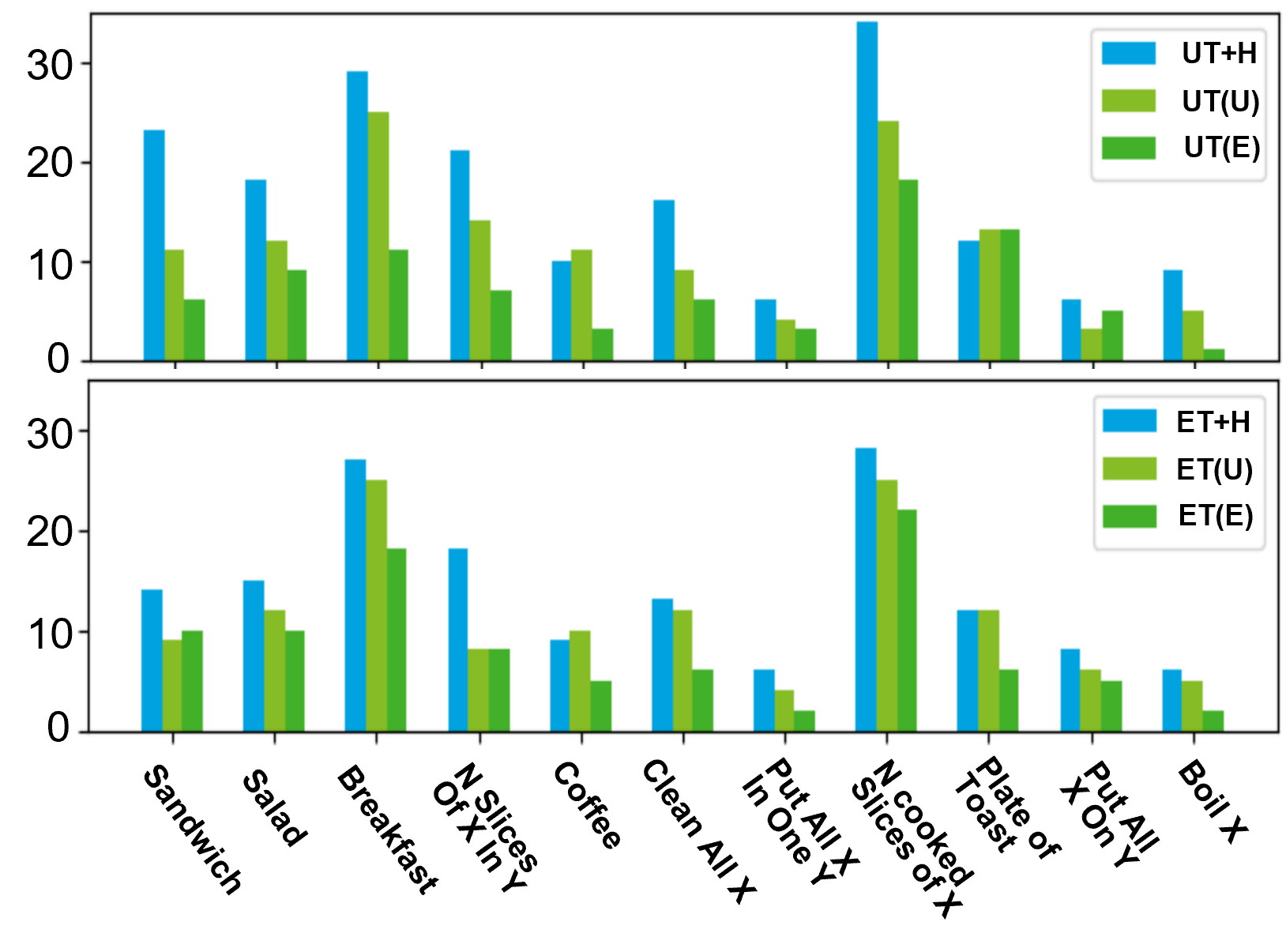}
    \caption{\textbf{Histogram of successful tasks for different models on different types of tasks.} The upper figure shows the UT model results, and the lower figure shows the ET model results.Blue column represents the hybrid training strategy, the light green represents training with the unit grained data, and the dark green represents  training with EDH segmentation data.}
    \label{successful task}
    \vspace{0em}
    \end{figure}
    
    \subsection{Experiment Setup}

    \paragraph{Datasets}
    We use the EDH benchmark from TEACH dataset\cite{padmakumar2021teach}, which is split into three parts: train, valid-seen and valid-unseen. Our unit segmentation instances are collected in the train split and we use them to train our model. All models and baselines are evaluated on the valid-seen and valid-unseen split of the EDH benchmark

    \paragraph{Evaluation Metrics}
    We evaluate our model using the evaluation metrics of the EDH dataset in TEACH. The Metrics including 4 parts: (1) success rate (\textbf{SR}) evaluates wether agent complete the task successfully; (2) goal-condition success (\textbf{GC}) evaluates the progress of the agent in completing the task ; (3) path weighted success rate (\textbf{PSR}) and path weighted goal condition success (\textbf{PGC}) are SR and GC weighted by the path length, which are used to evaluate the efficiency of the agent to complete the task . 

    \paragraph{Comparison Models}
    (1) \textbf{Seq2Seq}(Seq)~\cite{shridhar2020alfred} uses the previous hidden state and text output of the LSTM for attention, concatenating the representation of current image and previous action to predict the next action.
    (2) \textbf{Episodic Transformer}(ET)~\cite{pashevich2021episodic,padmakumar2021teach} takes all historical pictures and historical action information as input, and uses the current image representation after feature fusion to predict the next action.
    (3) \textbf{Unit Transformer}(UT) is our proposed model introduced in section \ref{unit-model}.
    (4) \textbf{Seq2seq with hybrid}(Seq+H) is Seq2seq model trained by hybrid forcing training framework.
    (5) \textbf{Episodic Transformer with hybrid}(ET+H) is UT trained by hybrid forcing training framework.
    (6) \textbf{Unit Transformer with hybrid}(UT+H) is UT trained by hybrid forcing training framework.

    \paragraph{Implementation Details}
    Object labels and region features are extracted from a trained Faster-RCNN \cite{pashevich2021episodic}. The sequential relationship between units from the same TEACH session makes parallel training not directly usable. To address this, we assign the same values for the unit initial state vectors and save these vectors as a global matrix. The global matrix is updated asynchronously via recording the final state vector of previous unit obtained during training as the initial state vector of the next unit. The training adopts a learning rate of 1e-3 with SGD optimizer.
    The random seed is fixed as 19980417 across all experiments.
    

    \begin{table}[t]
    \begin{center}
    \resizebox{\columnwidth}{!}{
    \begin{tabular}{lllll}
    \hline
    \multicolumn{1}{l}{\multirow{2}{*}{Model}} & \multicolumn{2}{c}{val-seen} & \multicolumn{2}{c}{val-unseen} \\ \cline{2-5} 
    \multicolumn{1}{l}{} & SR(PSR) & GC(PGC) & SR(PSR) & GC(PGC) \\ \hline
    Seq(E) & 0.8(0.2) & 1.5(0.9) & 4.4(1.4) & 5.3(4.6) \\
    Seq(U) & 2.1(0.9) & 2.6(2.0) & 5.1(1.7) & 5.9(5.0) \\
    \hline
    ET(E) & 4.5(0.7) & 4.4(2.4) & 6.0(1.6) & 5.0(4.8) \\
    ET(U) & 5.1(1.9) & 4.9(3.1) & 6.3(1.8) & 6.4(5.2) \\
    \hline
    UT(E) & 3.8(1.5) & 3.9(3.1) & 5.5(1.6) & 6.0(5.9) \\
    UT(U) & 6.8(2.0) & 6.6(3.9) & 7.4(2.4) & 7.2(7.4) \\
    \hline
    UT+H & \textbf{8.4(2.6)} & \textbf{6.8(6.1)} & \textbf{9.1(3.0)} & \textbf{9.4(9.5)} \\ \hline
    \end{tabular}
    }
    \end{center}
    \caption{\textbf{Main experiment results.} We compare the proposed Unit Transformer with seq2seq model and previous state-of-the-art ET model. The brackets of the model name indicate the segmentation method of the data used for training the model, E indicates EDH segmentation, and U indicates unit segmentation.}
    \label{main_result_table}
    \end{table}

    \subsection{Main Results}

    We train three models (Seq2Seq model, Episodic Transformer, and our proposed Unit Transformer) using both the original EDH benchmark training set and our new unit-grained training set. We evaluate these models on both seen and unseen validation datasets from the EDH benchmark and present the results in Table \ref{main_result_table}. The characters in parentheses after each model name indicate the type of data segmentation used during training (E for EDH segmentation and U for unit segmentation). Firstly, we observe that models trained with unit-grained instances outperformed those trained with EDH instances. Secondly, our proposed UT model increased the success rate by 35\% on the unseen validation set when trained with unit-grained data. The other two models did not show as significant an improvement with unit-grained training, indicating that our model is particularly well-suited for this type of training. Thirdly, when using unit-grained data for training, adding our proposed hybrid training strategy improved the performance of our model by another 22\%, providing evidence that our hybrid training strategy is highly effective.
    

    \begin{figure*}[h]
    \centering
    \setlength{\abovecaptionskip}{0.1cm}
    \includegraphics[width=\linewidth]{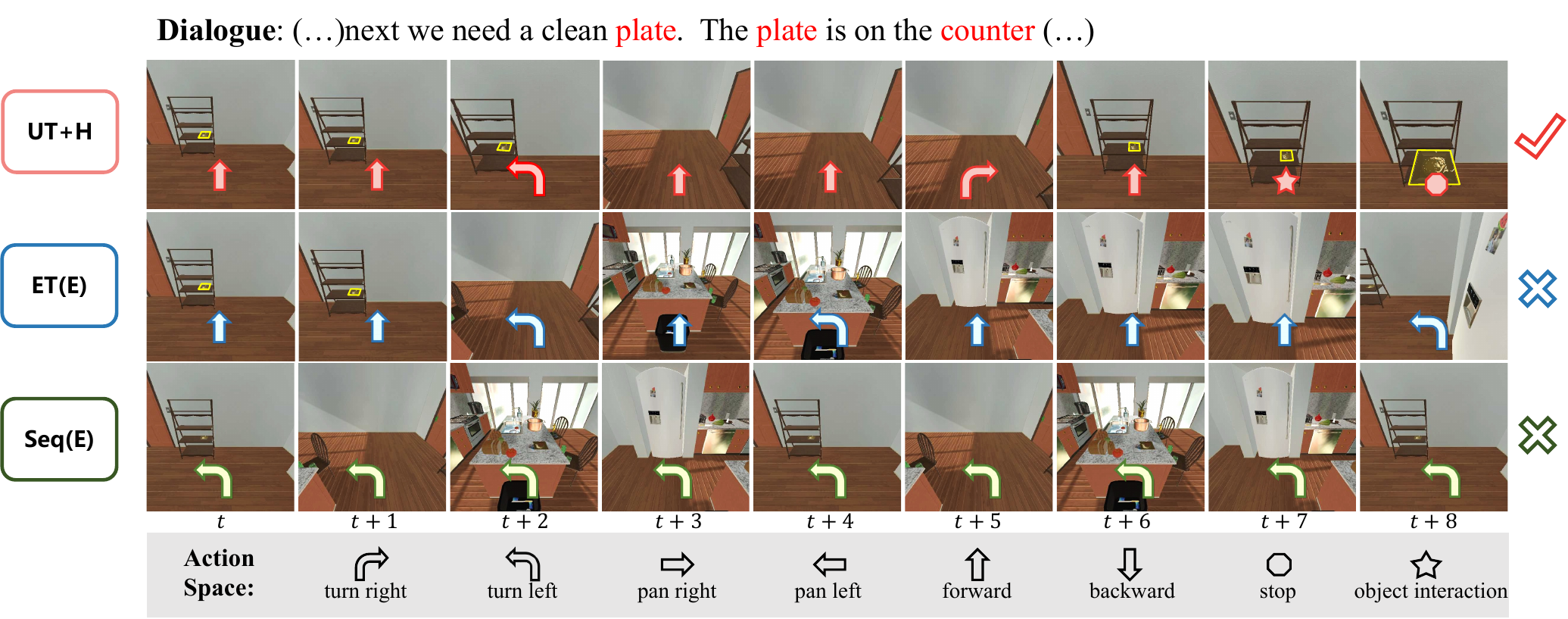}
    \caption{\textbf{Analysis of the agent performance in the same unit.} The yellow polygon in the picture highlight the location of the target, respectively. The red, blue and green action sequence is made by UT+H, ET(E) and Seq(E).}
    \label{visualization}
    \vspace{0em}
    \end{figure*}
    
    \subsection{Effectiveness of Hybrid Training}
    To investigate the generalizability of our proposed hybrid forcing training strategy, we apply it to two additional models, the Seq2Seq and ET models, and evaluate their performance on the EDH benchmark dataset. The experimental results are presented in Table \ref{hybrid training exp}. Comparing the results in Table \ref{main_result_table} and Table \ref{hybrid training exp}, we observe that the Seq2Seq, ET, and UT models all exhibit improved performance on both seen and unseen split under hybrid training strategy. Notably, the success rate of the Seq2Seq model on the seen validation set increased from 2.1\% to 6.8\% with the use of the hybrid forcing training strategy, demonstrating significant performance gains even for relatively simple models. Furthermore, we find that the path length weighted metrics of all models improves after incorporating the hybrid training strategy, suggesting that such a strategy enhances the trajectory fidelity. We think the improvement results from the reduced gap between training and inference.
    
    \begin{table}[t]
    \begin{center}
    \resizebox{\columnwidth}{!}{
    \begin{tabular}{cllll}
    \hline
    \multicolumn{1}{l}{\multirow{2}{*}{Model}} & \multicolumn{2}{c}{val-seen} & \multicolumn{2}{c}{val-unseen} \\ \cline{2-5} 
    \multicolumn{1}{l}{} & SR(PSR) & GC(PGC) & SR(PSR) & GC(PGC) \\ \hline
    Seq+H & 6.4(1.5) & 4.7(3.2) & 6.4(1.7) & 6.5(6.4) \\
    ET+H &  6.7(2.1)& 6.4(2.8) & 7.5(3.1) & 6.5(8.7)  \\
    UT+H & \textbf{8.4(2.6)} & \textbf{6.8(6.1)} & \textbf{9.1(3.0)} & \textbf{9.4(9.5)}\\ \hline
    \end{tabular}
    }
    \end{center}
    \caption{\textbf{Result of hybrid training strategy effectiveness experiment.} The performance of all three models has improved after using the hybrid training strategy}
    \label{hybrid training exp}
    \end{table}

    \begin{table}[t]
    \begin{center}
    \resizebox{\columnwidth}{!}{
    \begin{tabular}{lllll}
    \hline
    \multirow{2}{*}{Model} & \multicolumn{2}{c}{val-seen} & \multicolumn{2}{c}{val-unseen} \\ \cline{2-5} 
     & SR(PSR) & GC(PGC) & SR(PSR) & GC(PGC) \\ \hline
    UT & \textbf{6.8(2.0)} & \textbf{6.6(3.9)} & \textbf{7.4(2.4)} & \textbf{7.2(7.4)} \\
    -r & 4.1(1.7) & 3.3(3.3) & 4.7(1.6) & 5.2(6.5) \\
    -m & 6.2(1.7) & 6.5(5.3) & 5.3(1.3) & 5.6(4.8) \\
    -m-r & 3.3(1.6) & 3.0(3.2) & 4.0(1.8) & 5.5(7.0) \\ \hline
    \end{tabular}
    }
    \end{center}
    \caption{\textbf{Ablation experiment results of UT model}. "-r`` means to remove region feature of object, "-m`` means to remove the memory state feature, "-m-r`` means to remove both features.}
    \label{ablation study}
    \end{table}
    
    \subsection{Ablation Studies}
    In our proposed model, we introduce the object region feature and state memory vector as additional information. Table \ref{ablation study} explores the impact of these features on the performance of the UT model. The results indicate that removing either the object region feature or the memory state vector independently leads to a decrease in model performance. When the object region feature is removed, the success rate on both the seen and unseen split is reduced by 40\% and 36\%, respectively. Conversely, when the memory state vector is removed, the success rate on the seen and unseen split only drops by 8\% and 28\%, respectively. These results suggest that object information is more critical than memory state in the VLDM task, as there are numerous actions that require the identification and interaction with objects.
    
    \subsection{Analysis of Successful Tasks }
    We investigate the impact of utilizing different data granularity and training strategies on the success rate across different types of tasks. As shown in Figure \ref{successful task}, statistical results indidate that
    models trained with unit-grained data by a hybrid training strategy significantly surpass the performance of others when faced with complex tasks involving multiple objects and longer action sequences, such as making sandwiches or masking breakfast. These challenging tasks require the advanced agent ability to interact with multiple objects, making the unit-grained data segmentation and hybrid training strategy particularly effective.
    
    \section{Qualitative Analysis}
    A qualitative example is shown in Figure \ref{visualization}. In this scenario, the dialogue instructs the agent to navigate and pick up an empty plate. The proposed Unit Transformer, utilizing a hybrid training strategy, successfully navigates to an empty plate and then picks up the plate on the counter top as directed by the hint while other two baseline models either fails to find the plate or becomes trapped in a loop. This demonstrates the effectiveness of our method in navigating to objects specified in dialogue and interacting with them. Furthermore, utilizing the hybrid training strategy prevents the agent from getting caught in a loop during inference.
    

\section{Conclusion}
In this work, we propose a novel unit-grained instance segmentation method that enables agents to learn better by effectively segmenting data into smaller, more manageable units. Using this approach, we create an offline environment for each unit by collecting panoramas of every reachable point in each scene. We also introduce a hybrid training strategy that involves student forcing training and teacher forcing training, which reduces the gap between the training and inference process. Our experimental results demonstrate that this strategy can significantly improve performance when agents face more complex tasks. We also propose a Unit Transformer model that inputs image features of objects and uses a memory state vector to record historical information between different units. Through experiments, we validate that our proposed unit-grained instances and hybrid forcing training strategy is model-agnostic and can significantly improve the agent performance on vision-based tasks. Overall, our work presents a promising approach for vision-based agents by utilizing unit-grained data segmentation and hybrid training strategies. Future research could explore the effectiveness of these methods on other tasks and datasets and further investigate their generalizability.


{\small

\bibliographystyle{ieee_fullname}
\bibliography{costom}
}

\end{document}